# A Memory-Augmented LLM-Driven Method for Autonomous Merging of 3D Printing Work Orders

Yuhao Liu; Maolin Yang; Pingyu Jiang*

*Abstract*—With the rapid development of 3D printing, the demand for personalized and customized production on the manufacturing line is steadily increasing. Efficient merging of printing workpieces can significantly enhance the processing efficiency of the production line. Addressing the challenge, a Large Language Model (LLM)-driven method is established in this paper for the autonomous merging of 3D printing work orders, integrated with a memory-augmented learning strategy. In industrial scenarios, both device and order features are modeled into LLM-readable natural language prompt templates, and develop an order-device matching tool along with a merging interference checking module. By incorporating a self-memory learning strategy, an intelligent agent for autonomous order merging is constructed, resulting in improved accuracy and precision in order allocation. The proposed method effectively leverages the strengths of LLMs in industrial applications while reducing hallucination.

## I. INTRODUCTION

3D printing is an advanced manufacturing technology based on additive principles such as fused deposition modeling (FDM), stereo lithography apparatus (SLA), and selective laser sintering (SLS), where materials are deposited layer by layer to form parts. It enables the direct fabrication of complex-shaped components from 3D digital models without the need for complicated manufacturing processes. Due to its simplified and highly integrated workflow, along with its ability to support diversified, small-batch, and highly customized production, 3D printing has demonstrated significant advantages in terms of production flexibility and responsiveness [1,2]. Efficient allocation and merging of work orders can substantially improve production efficiency.

In response to this gap, this study presents a novel autonomous order merging method for 3D printing, which leverages a large language model (LLM) augmented with memory-enhanced learning capabilities. The method models the functional, performance, and structural features of both printing devices and work orders, transforming them into standardized prompt templates readable by LLMs. Based on these templates, an order-device matching prediction tool and an order merging interference checking tool are developed. With the aid of a multimodal large language model, autonomous decision-making and dynamic adjustment are achieved. Furthermore, by integrating a memory-based learning strategy, the system accumulates decision-making experience through iterations, continuously improving its accuracy and reliability. This paper introduces the work through four main sections: related work, methodology, case validation, and conclusion.

## II. RELATED WORK

The autonomy and merging problem of 3D printing orders essentially aims to solve the resource allocation and coordination optimization among multiple processing tasks. It focuses on the scheduling problem in a production workshop composed of multiple 3D printers. The characteristics of 3D printing must be considered, namely that the workpieces are produced in single or small batches, and each device can simultaneously produce multiple workpieces. The number of workpieces produced depends on the size of the printer and the size of the workpieces being printed. Therefore, the research on the autonomy and merging of 3D printing orders mainly includes two aspects: first, assigning production tasks to each production device; second, the placement problem of different workpieces within each printer.

Rational assignment of work orders and operations can significantly reduce production costs. Printing different types of parts simultaneously can greatly decrease unit product costs [3]. However, to group parts into a single batch for production, it is first necessary to resolve the placement problem of parts within a batch—ensuring that there is no interference between parts and that the printer's available space is fully utilized. Traditional approaches to autonomy and merging in 3D printing production lines mainly consist of two steps [4]: First, during production scheduling, parts and devices are classified based on their dimensions. Parts awaiting processing are then allocated to different devices according to these classifications, and the production batches for each device are determined. Second, third-party nesting software is used to determine the specific placement of each part within the production batch. Although classifying by type simplifies the problem and enables rapid allocation, the classification criteria largely depend on accumulated production experience, and the allocation outcome is highly reliant on managerial expertise.

Current research on autonomous merging of 3D printing orders mainly focuses on two types of spatial merging: two-dimensional and three-dimensional. Zhang, et al. [5, 6] combined two-dimensional irregular polygon optimization methods with genetic algorithms to optimize the 3D printing process in 2D space, aiming to achieve more compact part

*Research supported by the National Natural Science Foundation of China (No. 52375512).

Y. Liu. Author is a doctoral student at the State Key Laboratory for Manufacturing Systems Engineering, Xi'an Jiaotong University, Xi'an 710049, China. (e-mail: yuhaoliu@stu.xjtu.edu.cn).

M. Yang. Author is an assistant professor at the State Key Laboratory for Manufacturing Systems Engineering, Xi'an Jiaotong University, Xi'an 710049, China. (e-mail: maolin@xjtu.edu.cn).

P. Jiang.* Author is a professor at the State Key Laboratory for Manufacturing Systems Engineering, Xi'an Jiaotong University, Xi'an 710049, China. ( corresponding author to provide e-mail: pjiang@mail.xjtu.edu.cn).

placement and improve the utilization of the printer's workspace. Canellidis, et al. [7, 8] applied the No-Fit Polygon algorithm along with evolutionary algorithms to optimize SLA process. Gogate and Pande [9] conducted preliminary research on three-dimensional optimization and achieved promising results. Wu, et al. [10] further employed multi-objective optimization methods to tackle the 3D bin-packing problem in 3D printing.

However, these studies primarily focus on individual machines and aim to improve the productivity of a single device. They do not consider coordination among multiple machines, which can lead to workload imbalance and bottleneck stations in workshops consisting of multiple printers.

At present, most research on autonomous merging of 3D printing orders still centers on batch division for individual machines. There is relatively limited research addressing scenarios involving simultaneous production across multiple devices. Such multi-device scheduling problems are more complex and lack sufficient studies on holistic optimization methods.

III. METHOD

To address the current challenges in autonomous control and planning of 3D printing work orders, this paper proposes an LLM-based autonomous work order merging method driven by a multimodal large model with an integrated memory-augmented learning strategy. As illustrated in Fig 1, the proposed framework consists of a printing autonomy decision layer and a printing execution layer. It begins with production line modeling, in which the functional, performance, and structural characteristics of the 3D printing line are modeled. Based on this, a set of callable tools including an intelligent order matching tool and an interference checking tool is developed to support the construction of autonomous agents.

Moreover, the framework adopts a memory-augmented learning strategy, allowing the agent to accumulate successful cases and decision-making experience during each autonomous operation. Over time, this leads to continuous improvements in operational efficiency and decision accuracy.

During the autonomous work order process, the Printing Autonomy Decision Layer collects the operational status of each printer via IoT-enabled terminal devices (e.g., Raspberry Pi), aggregating this information to form a comprehensive view of the production line state. Meanwhile, the agent continuously receives new work orders and autonomously performs real-time processing, including order matching and merging tasks. Additionally, the agent supports human–machine interaction, allowing operators to monitor and intervene in the agent's decision-making process as needed.

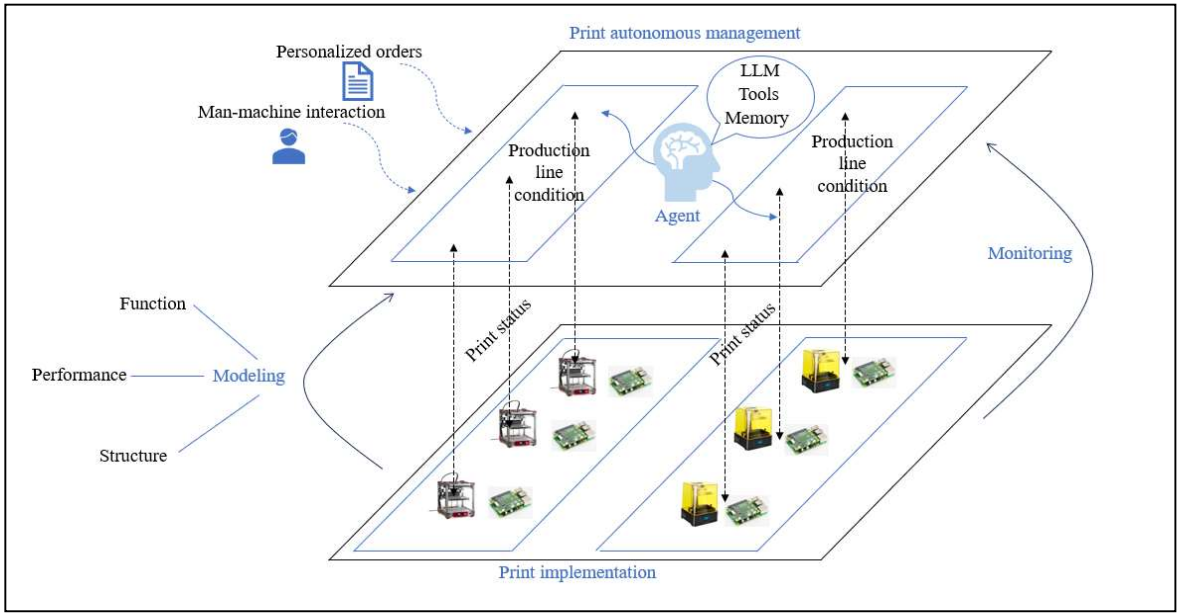

Figure 1.  Framework of Memory-Augmented LLM-Driven Autonomous Merging Method for 3D Printing Work Orders

## A. Feature Modeling of Printers and Work Orders

First, it is necessary to model the features of each 3D printer, which include functional, performance, and structural attributes. The functional features describe the printing capabilities of the printer, such as FDM, SLA, and other supported technologies. These capabilities and their corresponding printer mappings are represented in the form of sets, as shown below. Once the modeling is completed, the information is stored in JSON format for structured access and downstream use.

$$M = \{m_1, m_2, ......m_n\} \quad (1)$$

The set $M$ represents the collection of devices in the production line, where each element $m$ denotes an individual device.

$$m_n = \{FU, PR, ST\} \quad (2)$$

The element m represents a 3D printer, and its associated attributes are defined as follows: $FU$ denotes the functional features, $PR$ denotes the performance features, and $ST$ denotes the structural features of the printer.

$$FU = ['M\_type', 'Material'] \quad (3)$$

$FU$ is used to describe the type of the 3D printer, including the printing technology (e.g., FDM, SLA) and the types of materials it supports (e.g., PLA, ABS, etc.).

$$PR = ['M\_accuracy', 'M\_speed'] \quad (4)$$

$PR$ is used to describe the performance of the 3D printer, including attributes such as printing accuracy, printing speed, and other relevant performance indicators.

$$ST = [L, W, H] \quad (5)$$

$ST$ represents the structural description of the 3D printer, where $L$, $W$, and $H$ denote the length, width, and height of the printer's build volume, respectively.

Next, it is necessary to define the feature representation of work orders:

$$OR = \{or_1, or_2, ......or_n\} \quad (6)$$

$OR$ represents the set of work orders, where each element $or$ denotes an individual work order.

$$or_n = \{RA, MA, PR, ST, ET\} \quad (7)$$

Each or contains the specific requirements of the work order, including: RA: the spatial requirement (e.g., size or bounding box of the printed part), MA: the material requirement (e.g., PLA, ABS), PR: the precision requirement, ST: the start time of the order, ET: the expected delivery time.

$$RA = [L, W, H] \quad (8)$$

$$MA = ['Material'] \quad (9)$$

$$PR = ['M\_accuracy'] \quad (10)$$

The modeling process can be carried out within an interactive Web application, either by manually inputting parameters or through human–machine interaction for automatic extraction. These structured data are then passed to the multi-modal large model in the form of prompt templates, enabling intelligent decision-making by the autonomous agent.

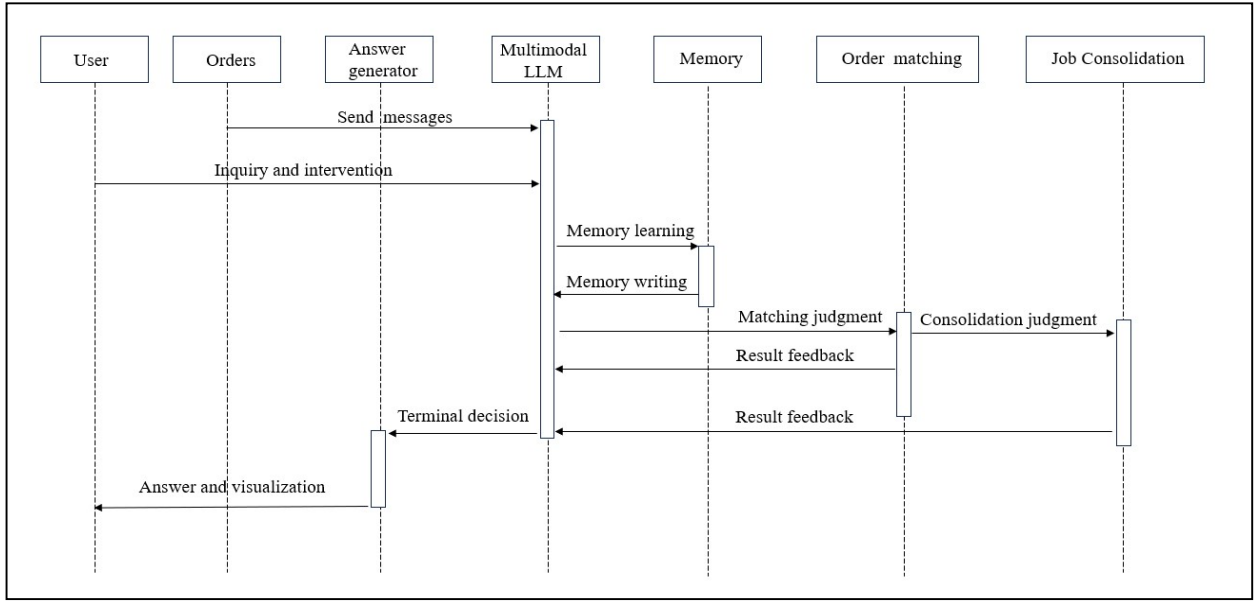

Figure 2. 3D printing order autonomous merging agent workflow

*B. Agent Tool Design*

In the context of autonomous work order merging, this study designs two key tools accessible to the agent: the Work Order Matching Tool and the Work Order Merging Interference Detection Tool. These tools enable the agent to autonomously retrieve, interpret, and apply decision-support information during task execution.

The Work Order Matching Tool evaluates the compatibility between work order requirements and the functional, performance, and structural characteristics of available 3D printers. It outputs a predicted set of optimal printer-order matches. The matching results are then passed to the LLM for reference in the decision-making process.

For interference detection, the proposed tool constructs individual bounding spaces around each part to be placed. An interference evaluation function is applied to detect spatial collisions between parts, as well as between each part and the printer's build volume.

Once the interference checks are completed, the results are returned in natural language format. These results are then passed to the LLM, which performs further reasoning and makes adaptive placement or merging decisions accordingly.

*C. Prompt Template Design*

In the operation of the autonomous agent, a critical component lies in leveraging the reasoning capability and accuracy of the LLM. Well-designed prompts are essential for reducing hallucinations and ensuring reliable responses. In this study, a prompt template was developed and refined through multiple rounds of experimentation and adjustment. The template is designed to meet the following requirements: role definition, parameterized structured input, and standardized output formatting.

For the work order merging process, an example prompt structure is illustrated in the Fig 3. This prompt includes the specific tasks to be performed by the LLM and integrates information obtained from the agent's tools (e.g., matching results, interference detection outcomes), prior successful cases, as well as images of merged layouts collected from each iteration. All of this information is passed to the multimodal large model to support informed, adaptive decision-making.

```
"role": "system",
"content":( "You are an expert in 3D printing layout optimization. "
    " Adjust object positions to avoid interference, "
    "ONLY changing x and y coordinates. Z coordinates must remain fixed at their initial values from "
    f"{default_positions}. Current positions are {positions}. The red box is the valid build volume "
        "(-50 to 50 in x and y, 0 to 100 in z). Objects are colored: 1.stl (red), 2.stl (green), 3.stl (blue). "
        "Use the interference report and images to suggest new x,y positions.
        Learn from these successful past layouts:"
    f"{memory_examples}"
        "Output in this format:"
        "```positions = [(x1, y1, z1), (x2, y2, z2), (x3, y3, z3)]```"
)
"role": "user",
"content": [
    {"type": "text",
    "text": f"Interference report:{interference_result}Adjust positions to resolve interference."},
    {"type": "image_url", "image_url": {"url": f"data:image/jpeg;base64,{view1_base64}"}},
    {"type": "image_url", "image_url": {"url": f"data:image/jpeg;base64,{view2_base64}"}},
    ]
```

Figure 3. Order merge autonomous prompt word template test example

## D. Agent Memory Learning and Management

As illustrated in Fig 2, within the agent's operational workflow, when a new work order is received by a specific production line, the relevant order information is transmitted to the multimodal intelligent agent in the form of messages. The agent first invokes the Work Order Matching Tool to autonomously predict and match the order with suitable processing equipment. Once the matching is completed, the agent then calls the Order Merging Tool to assess and compute the optimal placement of parts within the 3D printer based on the LLM's spatial reasoning.

The matching results and interference data from the merging process are fed back into the multimodal large model for further refinement, enabling it to adjust order allocation and part placement strategies dynamically.

In parallel, this framework incorporates a memory learning module. Throughout the continuous stream of order inputs and autonomous decisions, the agent selectively stores successful cases and strategies. When a new order arrives, the agent consults this memory to guide its initial decision-making, thereby improving the success rate and reducing the number of required iterations.

The decision outcomes are formatted into a standardized structure using the prompt template and returned to the human–machine interaction interface. Additionally, during agent operation, human operators can inject intervention instructions via the same prompt template, enabling the LLM to incorporate human-guided decisions when necessary.

## IV. CASE VERIFICATION

In this study, experiments were conducted using randomly generated work orders consisting of gear and rack parts. Material constraints were applied to the orders, and two manufacturing technologies, FDM and SLA, were selected. The proposed intelligent agent was used to autonomously assign work orders and arrange part merging for 3D printing tasks.

The detailed information of the work orders is shown in Table 1.

TABLE I. WORK ORDER DETAILS TABLE

| Order ID | Work order details | |
|---|---|---|
| | *Space / mm* | *Material* |
| CL01 | 24*23.99*10 | PLA |
| CL02 | 24*23.99*10 | PLA |
| CL03 | 24*23.99*10 | PLA |
| CT01 | 40.8*10*7 | EP Epoxy Resin |
| CT02 | 40.8*10*7 | EP Epoxy Resin |
| CT03 | 40.8*10*7 | EP Epoxy Resin |

The information of the devices is shown in Table 2.

TABLE II. DEVICE DETAILS TABLE

| Device | Device ID | Device details | |
|---|---|---|---|
| | | *Space/mm* | *Material* |
| 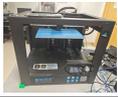 | EQ01 | 200*200*200 | PLA |
| 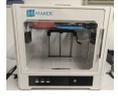 | EQ02 | 250*250*250 | PLA |
| 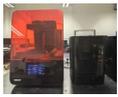 | EQ03 | 145*145*175 | EP Epoxy Resin |
| 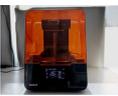 | EQ04 | 145*145*175 | EP Epoxy Resin |

Using the work order matching tool, the intelligent agent performed autonomous prediction and decision-making, assigning the corresponding orders to suitable 3D printers. Specifically, CL01, CL02, and CL03 were matched and merged for printing on EQ01, while CT01, CT02, and CT03 were assigned to EQ03 for merged printing.

After the intelligent agent made autonomous decisions, the orders were reasonably allocated based on printer types. For the gear orders, the intelligent agent assigned them to a single processing device for autonomous merging. After three iterations of decision-making, it successfully generated a correct work order merging and placement plan. The result is shown in Fig 4.

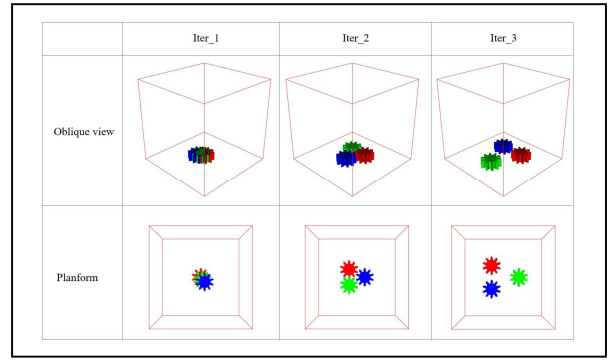

Figure 4. Gear Order Merging Iteration

For the rack orders, the agent, with the assistance of the interference detection tool, went through five iterations of autonomous decision-making before generating a correct and feasible work order merging plan. The result is illustrated in Fig 5.

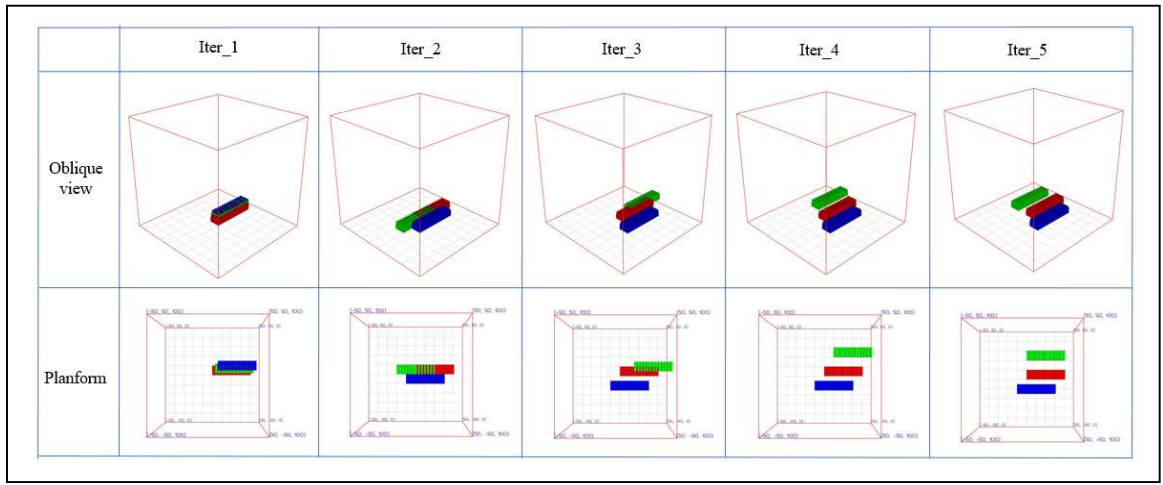

Figure 5. Rack Order Merging Iteration

## V. Conclusion

By modeling the features of production line equipment from the perspectives of functionality, performance, and structure, and integrating them with the modeling of work order characteristics, this study effectively conveys the autonomous status of production orders to the LLM through a well-designed prompt template framework. In addition, an intelligent agent framework was established, incorporating accurate judgment tools to support the LLM in reasoning and making informed decisions. Furthermore, a memory-augmented learning strategy was integrated into the iterative decision-making process, enabling the model to accumulate successful experiences based on its autonomous reasoning.

This research is of significant value for the development of personalized and autonomous order processing in 3D printing, enhancing the system's self-planning capability. The proposed framework also provides a reference for leveraging LLMs in industrial vertical applications, helping to reduce hallucination and improve autonomous planning performance. In future work, it will be necessary to model production execution data—such as Gantt chart-based monitoring and exception handling—and establish standardized prompt templates to enable the LLM to better understand the current situation and make accurate decisions.

## Acknowledgment

This work was supported by the National Natural Science Foundation of China (No. 52375512).